\documentclass{article}
\usepackage{graphicx} 
\usepackage{booktabs}
\usepackage{subcaption}
\usepackage{amsmath,amsfonts,amssymb}
\usepackage[hidelinks]{hyperref}
\usepackage{cleveref}
\usepackage{graphicx}
\usepackage{setspace}
\usepackage{tocloft}
\usepackage{url}

\usepackage{xcolor}
\usepackage{multirow}
	
\usepackage{lineno}
\usepackage{comment}
\usepackage{colortbl}
\usepackage{authblk}

\definecolor{LightCyan}{rgb}{0.88,1,1}
\definecolor{Gray}{gray}{0.7}

\title{Enhancing Post-Hoc Explanation Benchmark Reliability for Image Classification}
\author[1]{Tristan Gomez\thanks{tristan.gomez@univ-nantes.fr}}
\author[1]{Harold Mouchère\thanks{harold.mouchere@univ-nantes.fr}}
\affil[1]{Nantes Université, École Centrale Nantes, CNRS, LS2N, UMR 6004, F-44000 Nantes, France}

\date{July 2023}

\begin{document}

\maketitle

\begin{abstract}
    Deep neural networks, while powerful for image classification, often operate as "black boxes," complicating the understanding of their decision-making processes. Various explanation methods, particularly those generating saliency maps, aim to address this challenge. However, the inconsistency issues of faithfulness metrics hinder reliable benchmarking of explanation methods. This paper employs an approach inspired by psychometrics, utilizing Krippendorf's alpha to quantify the benchmark reliability of post-hoc methods in image classification. The study proposes model training modifications, including feeding perturbed samples and employing focal loss, to enhance robustness and calibration. Empirical evaluations demonstrate significant improvements in benchmark reliability across metrics, datasets, and post-hoc methods. This pioneering work establishes a foundation for more reliable evaluation practices in the realm of post-hoc explanation methods, emphasizing the importance of model robustness in the assessment process.
\end{abstract}

\section{Introduction}

I think it goes into a bit of detail too quickly. You should mention the following points:
- we're interested in the explanation for image models
- more particularly we focus on attribution map methods (like Grad-CAM)
- there are plenty of post-hoc methods and metrics to compare them.

Deep neural networks have emerged as powerful tools for various machine learning tasks, including image classification. 
However, their inherent complexity often renders them as black boxes, making it challenging to understand their decision-making process. 
To address this issue, numerous explanation methods have been developed.

In this work, we are interested in explaining image classification models and, more specifically, in generating attribution maps to shed light on the features used by the models.
Attribution maps can be generated in two main ways: (1) by embedding an attribution module within the model itself, constituting what is known as an attention model, or (2) through generic post-hoc methods applied to pre-trained models. 
To evaluate and compare attribution methods, the explainability community developped various metrics called \textit{faithfulness} metrics\footnote{In this paper, faithfulness metrics are also called attribution or fidelity metrics.}~\cite{AD,ADD,DAUC,DC}.
These metrics differ in their functionning but they all rely on a common principle: masking or highlighting parts of the input image and measuring if the prediction variation reflects the saliency of these regions, as estimated by the explanation.
By quantifying the faithfulness of explanations, one can construct benchmarks and compare attributions methods in terms of fidelity to the model's decision.

However, despite the extensive research on attribution methods and faithfulness metrics, a critical challenge remains: the lack of reliable benchmarking due to the inconsistency issues of these metrics.
One particular problematic aspect of these metrics is that the rankings they produce largely depends on the image on which the explanation is based.
Indeed, the metrics are unreliable to the point where they provide no insight on the studied post-hoc methods~\cite{sanity_checks}.
One potential explanation is that the samples generated to compute the saliency metrics are out of the training distribution (OOD) and lead to chaotic model behaviors.
This reliability issue is reminiscent of challenges faced in psychometrics, where tools have been developed to objectively assess the reliability of tests. 
One such tool is Krippendorf's alpha~\cite{Krippendorff}, a statistic designed to determine the degree of agreement among testers in a given evaluation context.

In particular, this statistic was already used by Tomsett et al. to evaluate the reliability of post-hoc method benchmarks~\cite{sanity_checks}.
Here, we reuse their protocol and employ it to compare the reliability of the benchmarks obtained with various training settings. 

The goal of this work is to improve the reliability of benchmarking post-hoc
methods for image classification. 
To do so, we propose several modifications to the training setting of models used for benchmarks and measure the reliability of the rankings obtained using Krippendorf’s alpha for each training setting. 

The first proposed modification consist in feeding the model with perturbed samples during training to improve robustness. 
More precisely, we study two types of perturbation: (1) perturbations obtained when computing the metrics to prevent OOD issues and (2) adversarial perturbations to smooth the model input space. 

Secondly, we replace the commonly used cross-entropy (CE) loss with the focal loss (FL) proposed by \cite{focal_loss_calib} to improve model calibration. 
The FL encourages the model to provide consistent confidence levels, such that increasing the masked proportion of an image, i.e. decreasing the accuracy, corresponds to a drop in confidence.

Through empirical evaluation, we demonstrate that these modifications and their combination can significantly improve the reliability of benchmarking post-hoc methods, as quantified by Krippendorf's alpha, depending on the fidelity metric, dataset and post-hoc methods considered.
For example, we show that the proposed modifications can reduce the number of images required to compute a post-hoc methods benchmark.
To the best of our knowledge, this is the first approach that attempts at improving the reliability of faithfulness metrics. 
This work emphasizes the importance of utilizing robust models when evaluating explanation methods, and we anticipate that future research on post-hoc methods will benefit from our proposed framework.

In summary, this paper aims to enhance the benchmark reliability of post-hoc methods for image classification by addressing the challenges associated with attribution metrics. 
By leveraging Krippendorf's alpha and introducing modifications to the model training process, we establish a more objective and robust framework for evaluating the performance of explanation methods.

\section{Related Work}

\subsection{Post-hoc methods}

Post-hoc methods can be divided into three groups:
\begin{itemize}
\item CAM-based methods. These methods compute a pondered mean of the last features maps of a convolutional neural network and differ by the method used to compute the weight of each feature map. The most popular CAM-based method is Grad-CAM (GC)~\cite{gradcam}.
\item Backpropagation-based methods. Given that CAM-based methods can only be applied on CNNs, backpropagation-based methods were developped to obtain explanation from any neural network by backpropagating information from the output to the input image. Popular examples include Guided Backpropagation (GP) and Integrated Gradients (IG).
\item Perturbation-based methods. Finally, to further expand the class of models that can be explained, perturbation-based methods were developped and consist in perturbing the input image to identify the area that contributes the most to a prediction. Popular examples include Occlusion and RISE.
\end{itemize}

\subsection{Faithfulness metrics \label{sec:faithfulness_metrics}}

As previously mentionned, the explainability community has developped many metrics to evaluate the faithfulness of explanations~\cite{captum,AD,ADD,DAUC,DC}.
In this work, we focus on the following metrics.
First, we use the Average Drop (AD) \cite{AD} and Average Drop in Deletion (ADD) \cite{ADD} metrics.
AD upscales the explanation and applies it as a mask on the input image to mask the least salient regions.
The metric is defined as the average drop in the model's confidence once the mask is applied.
ADD is similarly defined, except that the mask is reversed to highlight non-salient regions.

We also use the DAUC and IAUC metrics proposed by Petsiuk et al. \cite{DAUC}.
The DAUC metric perturbs the input image by accumulating black patches on the input image, starting from the least salient regions.
Then, the metric is defined as the area under the curve of the confidence/masked proportion curve.
The intuition is that the model's confidence should decrease as the masked proportion increases.
Similarly, the IAUC metric starts from a fully blurred image, and progressively removes the blur from the image.

We also use the variants proposed by Gomez et al. \cite{DC}.
The DC metric is a variant of the DAUC metric where the same perturbations are applied, but the metric is defined as the Pearson's correlation between the confidence varations and the saliency scores.
Similarly, the IC metric is a correlation-based variant of IAUC.

Finally, we also propose here non-cumulative variants of the DC and IC metrics.
Concretely, we do not accumulate the black patches on the image for DC and we do not progressively unblur the whole image for IC, and only reveal one patch at a time.
We name these variants respectively DC non-cumulative (DC-NC) and IC non-cumulative (IC-NC).
Note that one could also define DAUC-NC and IAUC-NC variants but such variants produce results that are independant of the chosen explanation.
Indeed, DAUC and IAUC evaluate explanations by the region ordering they produce, and if the perturbations are not accumulated, the order in which they are applied does not affect the AUC and the metric value becomes independant of the explanation.

\begin{figure}[ht]
    \centering
    \begin{subfigure}{0.4\textwidth}
        \centering
        \includegraphics[width=\textwidth]{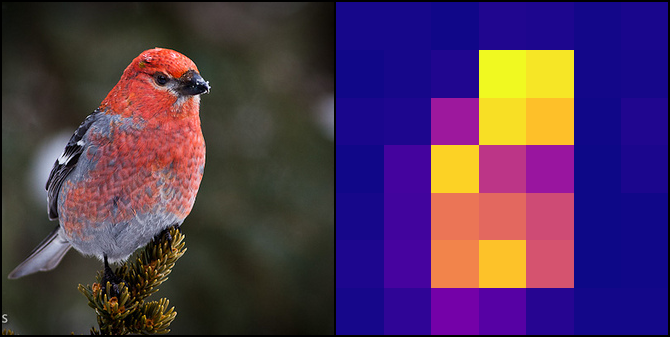}
        \caption{Original image and corresponding explanation.}
        \label{fig:metrics_samples_original}
    \end{subfigure}

    \begin{subfigure}{0.2\textwidth}
        \centering
        \includegraphics[width=\textwidth]{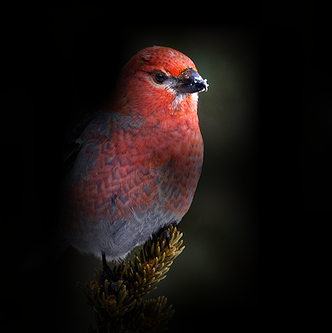}
        \caption{AD}
        \label{fig:metrics_samples_ad}
    \end{subfigure}
    \begin{subfigure}{0.2\textwidth}
        \centering
        \includegraphics[width=\textwidth]{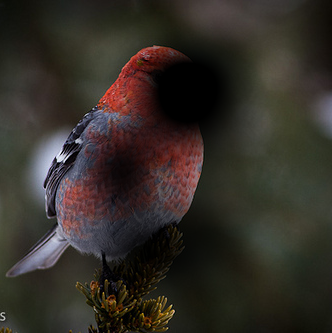}
        \caption{ADD}
        \label{fig:metrics_samples_add}
    \end{subfigure}
    \begin{subfigure}{0.2\textwidth}
        \centering
        \includegraphics[width=\textwidth]{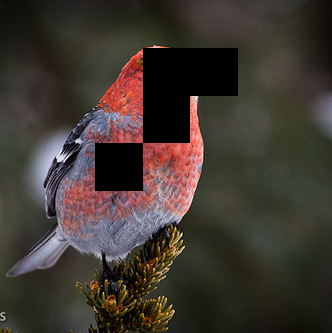}
        \caption{DAUC/DC}
        \label{fig:metrics_samples_dauc}
    \end{subfigure}
    \begin{subfigure}{0.2\textwidth}
        \centering
        \includegraphics[width=\textwidth]{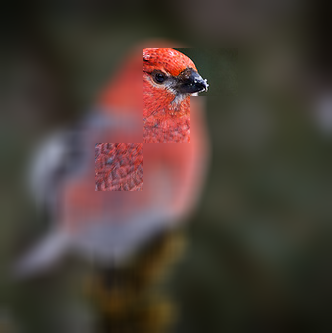}
        \caption{IAUC/IC}
        \label{fig:metrics_samples_iauc}
    \end{subfigure}
    \begin{subfigure}{0.2\textwidth}
        \centering
        \includegraphics[width=\textwidth]{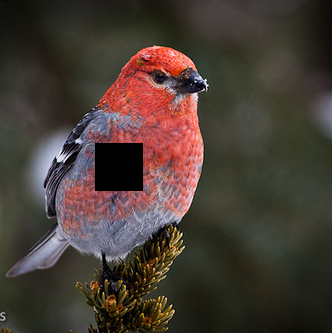}
        \caption{DC-NC}
        \label{fig:metrics_samples_dc}
    \end{subfigure}
    \begin{subfigure}{0.2\textwidth}
        \centering
        \includegraphics[width=\textwidth]{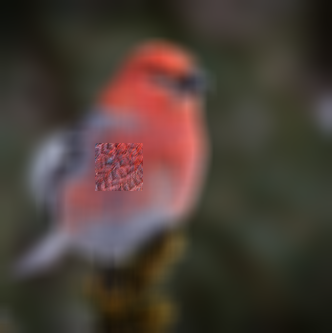}
        \caption{IC-NC}
        \label{fig:metrics_samples_ic}
    \end{subfigure}
    
    \caption{Examples of perturbations produced by the various metrics studied here.}
    \label{fig:metrics_samples}
\end{figure}

We show in \cref{fig:metrics_samples} examples of perturbation produced by the various metrics studied here.
Previous work showed that the perturbed samples are actually out of the training distribution, probably inducing a chaotic behavior of the model~\cite{DC}.
Hence, we propose to train the model on perturbed samples to prevent this issue.

\subsection{Krippendorf's alpha}

Tomsett et al.~\cite{sanity_checks} proposed the utilization of Krippendorf's alpha to assess the reliability of saliency metrics in image classification. 
Their study revealed significant inconsistencies in the rankings produced by these metrics. 
Building upon their work, we adopt their framework of using Krippendorf's alpha and propose modifications to the standard training setting to enhance benchmark reliability.

Krippendorf's alpha~\cite{Krippendorff} is a statistic developed by Krippendorf et al. to be applied in the psychometrics domain.
Following the analogy of Tomsett. et al., a collection of saliency methods can be seen as a battery of psychometric tests administered to the model. 

Each test is scored by each saliency metrics and each input image can be viewed as a distinct rater who administers the battery of tests.
Krippendorf's alpha then quantifies the degree of agreement among the raters, which in this case are the input images. 
Concretely, given a set of N test images, we generate M explanations for each image using different methods, resulting in a total of M explanations per image. 
Using a saliency metric, we assess the faithfulness of these explanations. 
By computing one explanation ranking for each image, we obtain a collection of N rankings. 

The level of agreement among these rankings is then quantified using Krippendorf's alpha.
The formula for calculating Krippendorf's alpha is as follows:

\[
\alpha = 1 - \frac{{D_o}}{{D_e}},
\]
where \(D_o\) is the observed disagreement, and \(D_e\) is the expected disagreement.
The observed disagreement \(D_o\) is the average proportion of disagreements or differences observed among the raters (i.e. images). 
It is computed based on the actual ratings or evaluations provided by the raters.
The expected disagreement \(D_e\) is the hypothetical disagreement that would occur by chance. 
It is calculated based on the distribution of explanation evaluations and takes into account the marginal distributions of the raters.

Perfect agreement is indicated by $\alpha=1$, while $\alpha=0$ indicates that the agreement similar to the one occuring by chance.
Note that $D_o$ can also be superior to $D_e$, resulting in negative values of $\alpha$.
Thus, Krippendorf's alpha ranges from $-\infty$ to 1.

\subsection{Focal loss}

In \cite{focal_loss_calib}, Mukhoti et al. proposed an adaptive variant of the FL to improve neural network calibration.
The authors show that the FL put less emphasis on easy samples, preventing the model from being overconfident on these samples.
The FL is defined as follows:

\[
\mathcal{L}_{focal} = - (1 - p)^\gamma \log(p),
\]
where $p$ is the model's confidence for the ground truth class and $\gamma$ is a hyperparameter.
The FL is a variant of the CE loss, where the term $(1 - p)^\gamma$ is multiplied to the loss.
As a consequence, if a sample is easy to classify, the model's confidence $p$ will be high and the loss will be low.
Mukhoti et al. provides a method to automatically compute the $\gamma$ parameter depending on the confidence on each sample and show that this adaptive FL variant significantly improves calibration compared to cross entropy.
Given its simplicity, we use this loss in our experiments to calibrate models.

\section{Method}

\subsection{Notations.}

Let $I\in \mathbb{R}^{H\times W\times3}$ be the image passed to the model, and $y_c$ be the score of the ground truth class.
We also note $(i,j) \in [0,...,H] \times [0,...,W] $ the spatial position on the input image and $(i',j') \in [0,...,H'] \times [0,...,W']$ the spatial position on the feature maps of the last layer.

\subsection{Training settings}

The goal of this work is to improve the reliability of benchmarking post-hoc methods for image classification.
To do so, we use Krippendorf's alpha not to compare faithfulness metrics like Tomsett et al.~\cite{sanity_checks} but to compare training settings.
As a consequence, the experiment highlights the importance of model training when benchmarking post-hoc methods.

The baseline model is trained on unmodified images with CE loss.
We compute perturbed image batches to be fed to the model along the regular batch to improve its behavior during the computation of faithfulness metrics.
We apply two types of perturbations. 
First, we generate faithfulness perturbations (FP) by applying a mask on the image computed using a faithfulness metric-inspired algorithm. 
This helps the model to learn how to process perturbations met during faithfulness metric computation and prevent OOD issues.
Secondly, adversarial perturbations (AP) are obtained with an adversarial attack method. 
The purpose is to smooth the input space of the model, which in turn could help yielding smoother and less noisy score curve when progressively masking the whole image, as with a multi-step metric for example.

We now define the following three training settings: 
\begin{itemize}
    \item FP: along the regular batch, a perturbed batch is computed by applying a mask that reproduce faithfulness metrics artifacts.
    \item AP: it is similar to FP, but the perturbations are computed using an adversarial attack method.
    \item FP+AP: combination of the first two perturbations. We compute one FP batch, one AP batch and a last one which first undergo FP and then AP (FP+AP).
\end{itemize} 

We use the same loss function for each batch and compute the final loss by simply adding together the individual losses. 
Finally, for each of the four aforementionned training settings (including the baseline), we also define a setting where we replace the CE function with the FL.

The purpose behind incorporating FL is to enhance the model's calibration, thereby potentially preventing the score from rising when progressively masking the entire image. 
Certainly, when progressively masking out an entire image, multi-step metrics expects the confidence score to decrease, but this may not necesseraly happen as usual standard training methods lead to poor calibration.

We train every possible combination of the proposed modifications.
In total, this amounts to 8 training settings that are summarized in \cref{tr_settings}.


\begin{table}[]
    \centering
    \resizebox{\textwidth}{!}{%
    \begin{tabular}{c|c|c|c|c|c}
        \toprule
        Name & Loss for all batches & Regular batch & FP batch & AP batch & FP+AP batch \\
        \midrule
        Baseline & CE & \checkmark &&& \\
        \rowcolor{Gray}
        FP & CE & \checkmark    & \checkmark && \\ 
        AP & CE &   \checkmark            && \checkmark &  \\
        \rowcolor{Gray} 
        FP+AP & CE & \checkmark & \checkmark & \checkmark & \checkmark \\
        \hline 
        FL & FL & \checkmark &&& \\
        \rowcolor{Gray}
        FL + FP & FL & \checkmark & \checkmark && \\ 
        FL + AP & FL &         \checkmark &     & \checkmark &  \\ 
        \rowcolor{Gray}
        FL + FP + AP & FL &\checkmark & \checkmark & \checkmark & \checkmark \\
        \bottomrule
    \end{tabular}%
    }
    \caption{Batches used for each training setting.}
    \label{tr_settings}
\end{table}  


We now describe how to compute an FP, an AP and an FP+AP batches.
Each image in a FP batch is computed using an explanation produced by a post-hoc method.
We note $I$, $I^{FP}$ and $S$ respectively the input image, the FP image and the saliency map explaining $I$.
For each image, we first randomly decide with equal probability which type of perturbation is going to be applied.
We distinguish four types of perturbations: 
\begin{itemize}
    \item AD-inspired: we apply the upsampled explanation as a mask on the image: 
    \begin{equation}
        I^{FP} = \textrm{norm}(\textrm{upsamp}(S)) \times I,
    \end{equation}
    where $\textrm{norm}(S)$ is the min-max normalization function, defined as $\textrm{norm}(S) = \frac{S-min(S)}{max(S)-min(S)}$, $\textrm{upsamp}(S)$ is a function that upsample S to the resolution of I, and $\times$ is the element-wise product. 
    \item ADD-inspired: we apply the reversed upscaled explanation as a mask on the image:
    \begin{equation}
        I^{FP} = (1-\textrm{norm}(\textrm{upsamp}(S))) \times I.
    \end{equation}
    \item DAUC-inspired: we select a random integer $k\in [0,H'W'-1]$, and mask the regions of the image corresponding to the top-$k$ pixels of the explanation with zeros. 
    Let $(i_k,j_k) \in [0,...,H'-1]\times[0,...,W'-1]$ be the spatial position of the top-$k$ pixel of the explanation, we define first a low-resolution mask as follows:
    \begin{equation}
        I^{FP-LR}_{ij} = 
    \begin{cases}
        0,& \text{if }\exists k' \in [1,k]~\text{s.t.}~ (i,j)=(i_{k'},j_{k'}) \\
        1,              & \text{otherwise,}.
    \end{cases}
    \end{equation}
    Then we upsample this mask with nearest-neighbor interpolation to the resolution of the input image:
    \begin{equation}
        I^{FP} = \textrm{upsamp}_{NN}(I^{FP-LR})
    \end{equation}
    \item IAUC-inspired: Similarly to DAUC-inspired, we sample an integer $k$ and blur the whole image except the top-$k$ areas.
\end{itemize} 
Note that to reduce computation, we obtain a rough saliency map $S$ by averaging the last feature maps of the CNN (post-ReLU).
Note that this is not one of the post-hoc methods we study, but we found it sufficient to obtain reliability improvements.

\begin{figure}
    \centering
    \includegraphics[width=\textwidth]{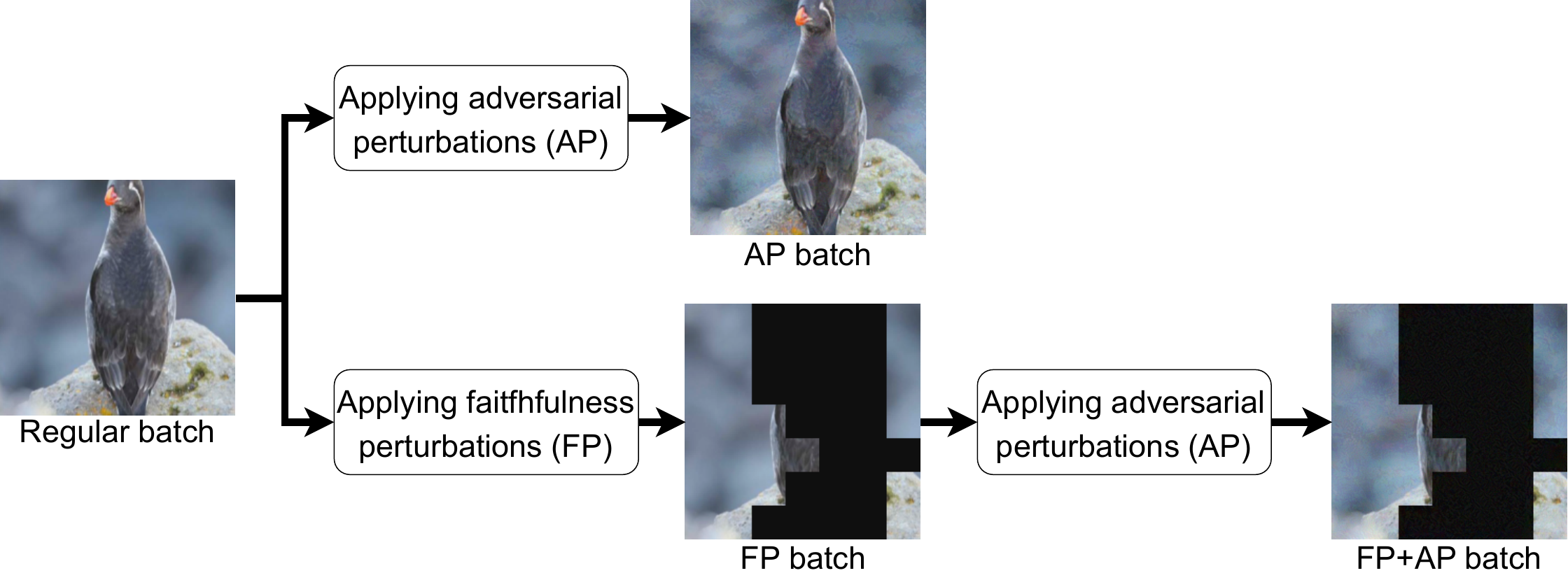}
    \caption{Illustration of the computation of the batches used for the various training settings.}
    \label{batchcomputation}
\end{figure}

The AP batch is computed by the usual adversarial training method: the regular batch is processed by the model, a perturbation is computed using an adversarial attack and is applied on the original images.  
In this work, the attack is computed using the Projected Gradient Descent (PGD)method~\cite{pgd}.
The FP+AP batch is obtained by computing an adversarial attack on the FP batch.
Batch computation is illustrated in \cref{batchcomputation}.

\subsection{Evaluation}

Once a model from each setting has been trained, we compute explanations using post-hoc methods.
More precisely, we use four method from each class of post-hoc methods from the litterature which amount to $M=12$ methods in total~\cite{tutorial}:

\begin{itemize}
    \item 4 CAM-based methods: CAM, Grad-CAM (GC), Grad-CAM++ (GC++) and Score-CAM (SC).
    \item 4 Backpropagation-based methods: Guided Backpropagation (GP), Integrated Gradients (IG), SmoothGrad (SG) and varGrad (VG).
    \item 4 Perturbation-based methods: Feature Permutation (FeatPerm), Feature Ablation (FeatAbl), Occlusion and RISE.
\end{itemize}

To compute the explanations, we use a set of N randomly chosen images from the test set.
To represent each class equally, we select C images from each class.
For each dataset, we select the smallest C such that the total number of images at least equal to 100.
For example the CUB dataset~\cite{CUB} has 200 classes, which means we must at least select 200 images (one from each class) to obtain a balanced representation of the classes and have at least 100 images.
On the other hand, the CROHNIPI~\cite{crohnipi} dataset only has 7 classes, which means we only need to select 15 images per class to obtain at least 100 images.

\begin{figure}[ht]
    \centering
    \includegraphics[width=\textwidth]{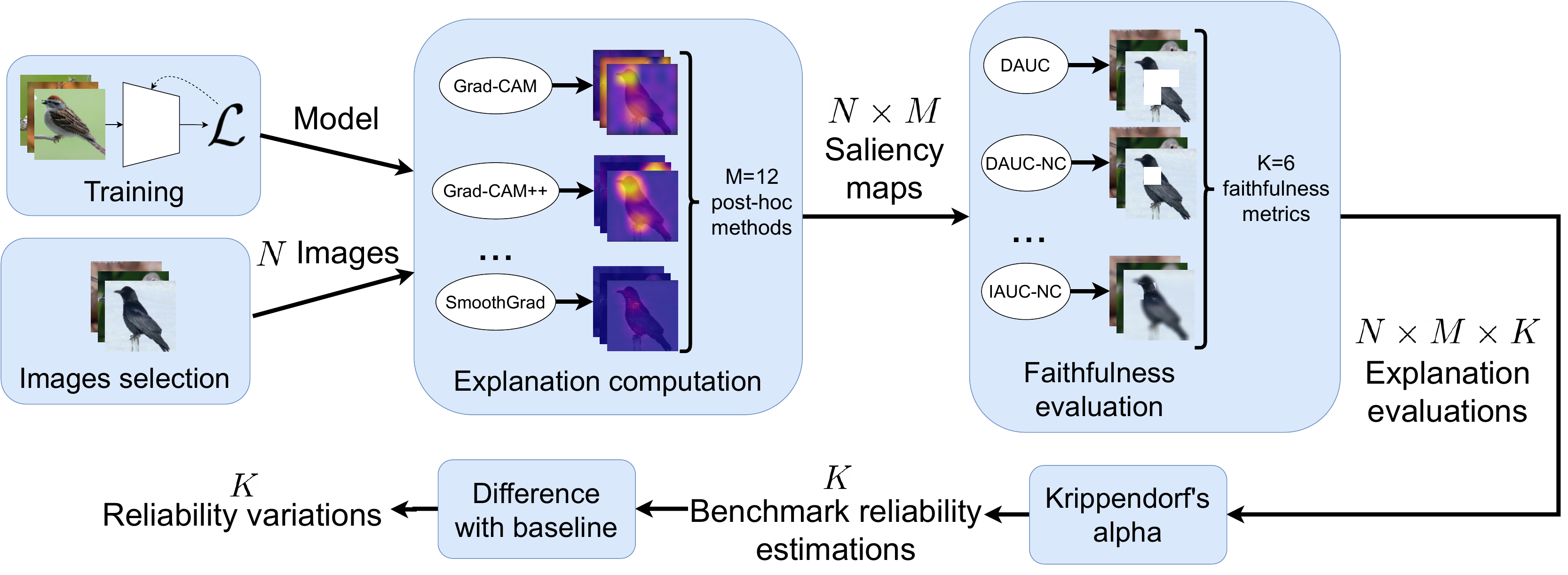}
    \caption{Workflow of the evaluation process.}
    \label{fig:workflow}
\end{figure}

Then, each explanation is evaluated by each one of the $K=8$ metrics described in \cref{sec:faithfulness_metrics}.
This produces K matrices of size ($N\times M$) containing the $N$ evaluations of the $M$ post-hoc methods.
Following previous work \cite{sanity_checks}, we convert each row of the matrices into a ranking (i.e. for each image, we sort methods according to their estimated fidelity), and compute Krippendorf's alpha on the rankings with the ordinal difference function to compare ranks~\cite{Krippendorff}.
We obtain $K$ krippendorf's alpha values for each training setting.
Finally we measure the alpha variation by comparing the $K$ values with the ones obtained with the baseline model.
If the variation is positive, is means that the modified training setting yield higher inter-raters aggreement, i.e. more consistent rankings across the $N$ test images, than the baseline setting.
On the other hand, if the variation is negative, it means that the modified setting generates even more inconsistent rankings than the standard setting.
This workflow is described in \cref{fig:workflow}.

\section{Experiments}

\subsection{Datasets}

First, we use the popular CUB-200-2011 (CUB)~\cite{CUB}, Standford Cars (CARS)~\cite{stanfordCars} and FGVC Aircraft (AIRCRAFT)~\cite{aircraft} datasets.
Secondly, we also employ two real-world fine-grained datasets. 
The CROHNIPI dataset~\cite{crohnipi} is a dataset of 3384 images of colonoscopy images belonging to 7 classes (6 pathological classes and 1 non-pathological class) and the Time-lapse embryo dataset~\cite{embryo_dataset}: a dataset of 59.831 microscopic images of human embryos belonging to 16 differents classes (the 16 development stages of the embryo).
\Cref{table:datasetinfos} shows the number of images per dataset and the number of images selected for the evaluation.

\begin{table}[]
    \centering
    \resizebox{\textwidth}{!}{%
    \begin{tabular}{c|c|c|c|c}
        \toprule
        \multirow{2}{*}{Dataset} & \multirow{2}{*}{Image number (training/validation+test)} & Class number & Selected image   & Total image number \\
                                 &                               &              & number per class & for evaluating explanations \\
        \midrule 
        CUB-200-2011 & 11788 (5994+5794) & 200 & 1&200 \\
        Stanford Cars & 16185 (8041+8144) & 196 &1 &196 \\
        FGVC Aircraft & 10000 (6667+3333) & 100 & 1&100 \\
        CROHNIPI & 3384 (2781+697) & 7 & 15 &105 \\
        Time-lapse embryo & 59831 (30104+29727) & 16 & 7 & 112\\    
        \bottomrule
    \end{tabular}%
    }
    \caption{Number of selected images per dataset for evaluating explanations.}
    \label{table:datasetinfos}
\end{table}

\subsection{Implementation details}

We use a ResNet-50 architecture for all experiments.
Hyper-parameters are shown in \cref{table:hyperparams}.

\begin{table}[]
    \centering
    \begin{tabular}{c|c}
        \toprule
        Hyper-parameter & Value \\
        \midrule 
        Batch size & 12 \\
        Learning rate & 0.001 \\
        Optimizer & SGD \\
        Momentum & 0.9 \\ 
        Weight decay & 1e-8 \\
        Classification layer dropout & 0.2 \\ 
        Epoch number & 100 (20 for embryo)\\ 
        Max worse epoch nb & 10 (5 for embryo)\\
        Input size & $224\times224$ \\
        \hline
        Saliency maps size & $7\times7$ \\ 
        RISE mask number & 8000 \\ 
        Noise tunnel samples & 30 \\
        \bottomrule
    \end{tabular}
    \caption{Hyper-parameters used for training the models. The total epoch number and max worse epoch number are lower for the embryo datasets as it is larger than the other datasets (60k vs. ~10k images).}
    \label{table:hyperparams}
\end{table}

When possible, we rely on Captum~\cite{captum} to compute all post-hoc methods except CAM and RISE, for which we implement our own version, and SC, for which we utilize Pytorch-GradCAM's implementation~\cite{pytorchcam}. 
We also employ an existing implementation for Krippendorf's alpha~\cite{krippendorf_impl} and leverage Scipy for bootstrap procedures. 
Models are trained with Pytorch and the code is available on Github."

To compute the adversarial attack we use the Projected Gradient Descent method~\cite{pgd} (PGD) with 4 steps, an alpha of $2/255$ and a maximum perturbation size of $1/255$.

\subsection{Results \label{results}}

In this section, we evaluate the reliability of post-hoc methods benchmarks using Krippendorf's alpha.
In particular, we measure the alpha variation between the baseline setting and each one of the 7 modified training settings.
To evaluate if the alpha variation is significant, we first compute alpha's distribution for every setting using bootstrap, i.e. by re-sampling the N rows of the matrices with replacement.
We generate $5000$ bootstrap samples to estimate the distribution.
Then, we first run a Shapiro test on the alpha distributions of the baseline and evaluated settings. 
If the test indicates that either of the distributions is not normal, we use a Mann-Whitney test to compare the two distributions. 
Otherwise, we use a Student's t-test. 
If we use a Student's t-test, we also run a Levene test to check the homogeneity of the variances.
For all tests, we compare the p-value to a 0.05 threshold.

\begin{figure}[ht]
    \centering
    \includegraphics[trim={2cm 0 2cm 0},clip,width=\textwidth]{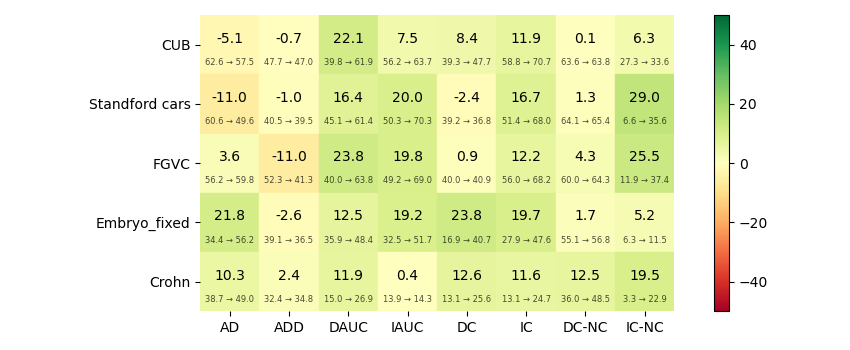}
    \caption{Variation of the mean of Krippendorf's alpha between the baseline setting and the FL+FP training setting for each metric and each dataset when ranking all the 12 post-hoc methods. In each square, the value in the center of the alpha variation. A positive value indicates an increase of Krippendorf's alpha. The values at the bottoms separated by ''$\rightarrow$`` show the baseline and modified setting alpha values. Values are multiplied by 100 for clarity. }
    \label{fig:alpha_variation}
\end{figure}

To illustrate the results obtained, we first show Krippendorf's alpha variation for the FP+FL training setting when ranking all the $12$ post-hoc methods in \cref{fig:alpha_variation}. 
All differences in this figure are statistically significant.

One can see that the impact of the FP+FL setting compared to the baseline is overall strongly positive but also depends on the metric and dataset considered.
For example, the increase obtained on the Standford cars dataset is 28.9 ($7.1 \rightarrow 36.0$) when considering the ADD metric whereas a 10.9 decrease ($52.7 \rightarrow 41.9$) is observed with the IC-NC metric.
On the other hand, when considering the AD metric on the Stanford cars dataset, Krippendorf's alpha decreases ($60.6 \rightarrow 49.6$).

For a better overall understanding of the impact of the various training settings used here, we compute the average alpha difference obtained across datasets and metrics for each training setting and each post-hoc class considered (Class-Map, Backpropagation (BP), and Perturbations).
We constitute another group of post-hoc methods with one explanation from each class: SC, RISE and SG, that we note $\{SC-RISE-SG\}$.
\Cref{avg_kripp_var} shows the mean variation of the krippendorf's alpha compared to the baseline performance depending on the post-hoc group considered.
The best training settings seems to be FP+FL, AP+FL and AP+FL, showing that the three training modification suggested in this paper seems to be useful.
However, the effect is largely dependant on the post-hoc group considered.
For example, FL seems to have a strongly positive effect when considering all the post-hoc methods, the perturbation-based methods or the $\{SC-RISE-SG\}$ group, but a negligible effect when considering the class-map-based or backpropagation-based methods.
More interestingly, combining modifications does not necessarily yield better results than the individual modifications by themselves, depending on the post-hoc group. 
This may imply that some modifications have a redundant effect or that their combination poses training issues.
We leave the study of such issues for future work.


\begin{table}[]
    \centering
    \resizebox{\textwidth}{!}{%
    \begin{tabular}{c|ccccc}\\ 
        \toprule 
        Model & ALL & Class-Map & BP & Perturbations & Recent\\
        \midrule 
        FP & 6.129 & 2.512 & 0.325 & 9.039 & 9.284\\
        AP & -1.262 & 4.777 & 3.920 & -0.839 & -3.166\\
        FL & 3.226 & 1.525 & 2.041 & 1.886 & 4.605\\
        \hline 
        FP+FL & 6.664 & 1.644 & 0.438 & \textbf{11.131} & \textbf{11.872}\\
        AP+FL & -0.771 & \textbf{7.160} & 4.589 & -1.309 & -3.073\\
        FP+AP & 4.831 & 3.816 & 5.212 & 5.188 & 4.353\\
        \hline 
        FP+AP+FL & \textbf{8.346} & 6.157 & \textbf{5.295} & 6.387 & 9.845\\
        \bottomrule 
        \end{tabular}
    }
    \caption{Mean krippendorf's alpha variation across datasets and metrics depending on the post-hoc group considered. A positive value indicates an increase of Krippendorf's alpha. Values are multiplied by 100 for clarity. \label{avg_kripp_var}}
\end{table}

\subsubsection{Impact on post-hoc benchmarks}

In this section we illustrate the impact of the proposed modifications on the pos-hoc methods benchmarks obtained.
More specifically, we focus on the benchmark of the 4 perturbation-based post-hoc methods using the DAUC metric on the CUB dataset and compare the baseline and the FL + FP settings.
Recall that we have $N$ test images on which we compute explanations which means that for one given metric, we  obtain $N$ rankings of the post-hoc methods.

\begin{figure}[ht]
    \centering
    \begin{subfigure}{\textwidth}
        \centering
        \includegraphics[width=\textwidth]{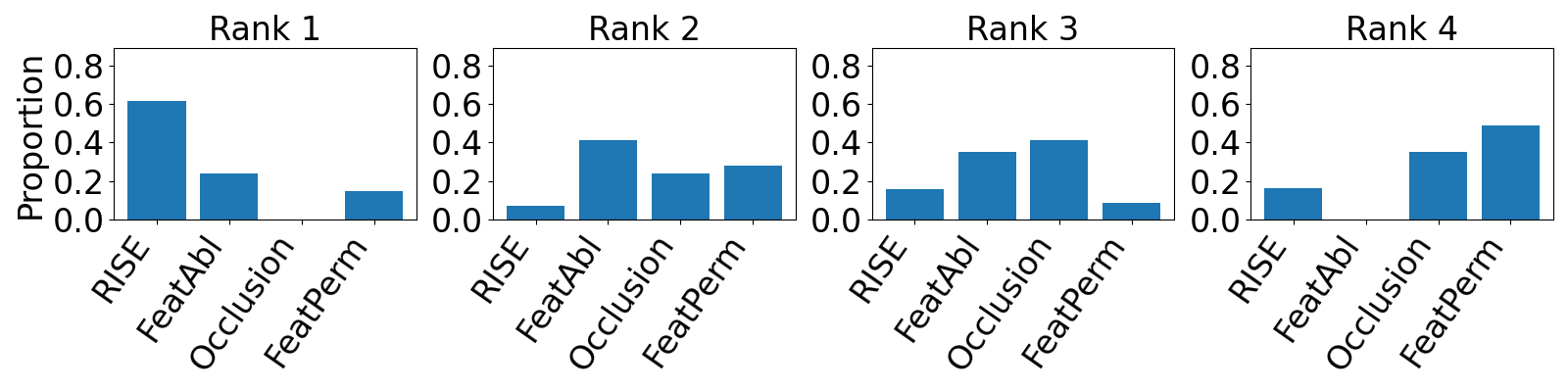}
        \caption{Baseline}
        \label{fig:rank_baseline}
    \end{subfigure}
    \begin{subfigure}{\textwidth}
        \centering
        \includegraphics[width=\textwidth]{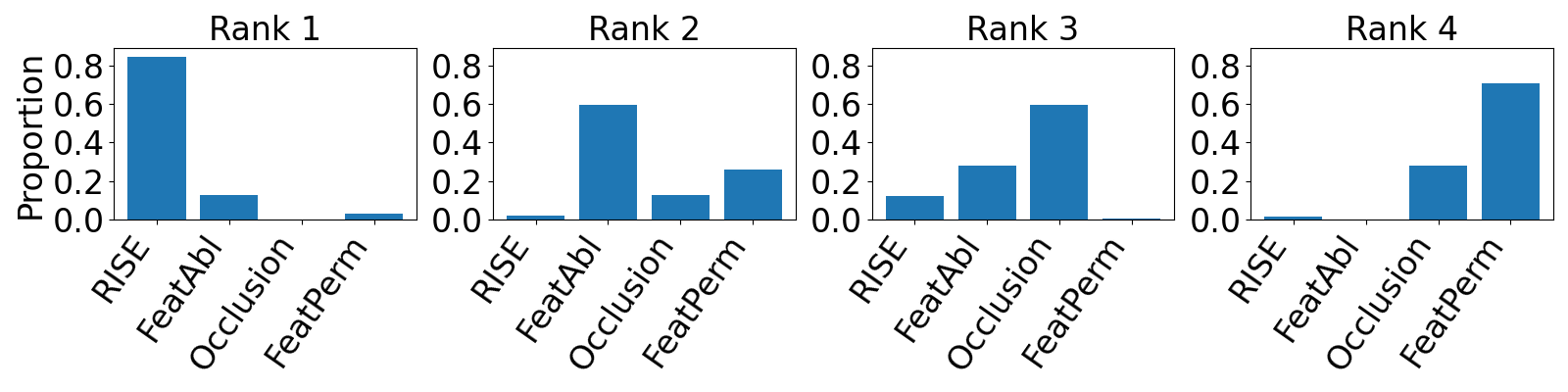}
        \caption{FL + FP}
        \label{fig:rank_pb_fl}
    \end{subfigure}    
    \caption{Distribution of rank position of the perturbation-based methods according to the DAUC metric.}
    \label{fig:rank}
\end{figure}

In \cref{fig:rank} we visualize the distribution of rankings for these methods, to determine which method tends to occupy each of the top-4 rank positions (in this case, there are four positions as there are four post-hoc methods in the perturbation-based group). 
To achieve this, we created histograms that illustrate the frequency with which each method achieved a particular rank position across the $N$ test images. 
For example, we can see in \cref{fig:rank_baseline} that among the $N$ ranking, approximately $60\%$ attribute the first place to the RISE method with the baseline training setting and \cref{fig:rank_pb_fl} shows that this number increases to more than $80\%$ with the FL+PB setting.
Similarly, with the baseline setting, it is ambiguous which is the top-3 performing method, whereas with the FL+PB setting, it is clearer that it would be Occlusion. 
More generally, the rankings of the FL+PB setting are more similar with each other than the baseline rankings, which allow to identify the following global top-4 ranking: RISE, FeatAbl, Occlusion and FeatPerm.

Using the training modifications proposed, the explanations rankings are more similar to each other, which means that one can reduce the number of images $N$ on which explanations are computed, while maintaining the same top-1 method, for example.
Note that reducing $N$ has significant benefits as the computation cost for post-hoc benchmarks is high. 
Indeed, simply computing the explanations has a high cost (particularly with perturbation-based explanations like RISE which require several thousand inferences per images), and computing a multi-step metric like DAUC or IAUC which scales quadratically with the explanation dimension size.
Hence, we study in the next section how much can the test size $N$ be reduced using the new training settings to further the analysis.

\subsection{Test size reduction}

In this section, we study how much can the number $N$ of images used to benchmark the post-hoc methods can be reduced while maintaining the same top-1 method.
We focus this analysis on the top-1 method for simplicity.
We note $n_i$ the number of images on which the post-hoc method $i$ was ranked first among all the $N$ images.
Let $i_{best}$ be the index of the post-hoc method identified as the best by the benchmark, i.e. such that:
\begin{equation}
n_{i_{best}} = \underset{i\in [1,M]}{max} \{n_i\}.
\end{equation}
 
Now, we want to identify the smallest $N' \leq N$ such that when sampling $N'$ images among the $N$ without replacement, the method $i_{best}$ stays the top-1 method.
We note $n'_i$ the number of images on which method $i$ is the best method among all the $N'$ images.
Formally, we are looking for the smallest $N'$ such that:
\begin{equation}
n'_{i_{best}} = \underset{i\in [1,M]}{max} \{n'_i\}
\end{equation}
Note that $n'_i$ is a random variable that depends on the subset of images selected.
Hence, we search for the smallest $N'$ such that the probability that method $i_{best}$ stays the best method is superior to $1-\alpha$ where $\alpha$ is a risk that we set to $0.05$:

\begin{equation}
P_{N'} = P(n'_{i_{best}} = \underset{i\in [1,M]}{max} \{n'_i\}) \geq 1-\alpha
\end{equation}

Therefore, we seek to solve the following problem:
\begin{equation}
N^* = \underset{N' \in [1,N]}{min} \left(P_{N'} \geq 1-\alpha \right), \label{problem}
\end{equation}
where we call $N^*$ the minimum benchmark size.

This probability can be computed by first listing all possibles outcomes for the counts $n'_i$.
For example if we benchmark $M=3$ methods on $N'=2$ images, the list of possible outcomes is shown in \cref{listpossibleoutcomes}.

\begin{table}[]
    \centering
    \begin{tabular}{ccc}\\ 
        \toprule 
        $n'_1$ & $n'_2$ & $n'_3$ \\ 
        \midrule 
        0 & 0 & 2 \\
        0 & 2 & 0 \\ 
        2 & 0 & 0 \\ 
        0 & 1 & 1 \\
        1 & 1 & 0 \\
        1 & 0 & 1 \\
        \bottomrule 
        \end{tabular}
    \caption{List of possible outcomes for the counts $n'_i$ when benchmarking $M=3$ methods on $N'=2$ images. \label{listpossibleoutcomes}}
\end{table}

Secondly, we compute the probability of each outcome using the probability mass function of the multinomial distribution:
\begin{equation}
    P(\bold{n'}_M) = 
\begin{cases}
    \frac{N'!}{n'_1\cdot n'_2\cdot ...\cdot n'_M} p_1^{n'_1} \cdot p_2^{n'_2} \cdot ... \cdot p_M^{n'_M} & \text{if} \sum_{i=0}^M n'_i = N'  \\
    0,              & \text{otherwise},
\end{cases}
\end{equation}

where $\bold{n'}_M$ is a shorthand for the tuple $(n'_1,n'_2,...,n'_M)$ and where we approximate the probabilities $p_i$ by the corresponding frequency on the full test set of size $N$: $p_i := f_i = n_i/N$.

Thirdly, we compute $P_{N'}$ by summing all the probabilities of outcomes where method $i_{max}$ is still ranked as the best method:
\begin{equation}
    P_{N'} = \sum\limits_{\bold{n'}_M \in Valid(N')} P(\bold{n'}_M), \label{sum}
\end{equation}
where $Valid(N')$ is the set of all $(\bold{n'}_M)$ tuples such that $n'_{i_{max}}>n'_i, \forall i \neq i_{max}$.

Note that the number of terms in \cref{sum} is combinatorialy large, which is why in practice \cref{sum} is approximated by a grid sampling of the tuples. 
More precisely, we define a proportion to sample $p_{samp}\in[0,1]$ among the range of values $[0,N']$ for each $n'_i$.
Then, when listing all possible tuples, we only evaluate those which satisfy the following condition:
\begin{equation}
n'_i~mod~1/p_{samp}=0, \forall i \in[1,M]
\end{equation}
We denote the ensemble of tuples that satisfy this condition $Sampled(N',p)$.
In the following experiment, we use $p=0.5$, which means that we only evaluate tuples with even counts.

Given that we do not evaluate all possible combinations, we update equation \cref{sum} by normalizing the sum as follows:

\begin{equation}
    P_{N'} = \frac{\sum\limits_{\bold{n'}_M \in Valid(N') \cap Sampled(N',p)} P(\bold{n'}_M)}{\sum\limits_{\bold{n'}_M \in Sampled(N',p)} P(\bold{n'}_M)}, \label{sum_norm}
\end{equation}

Now that we can compute $P_{N'}$, we solve \cref{problem} using binary search.
Indeed, $P_{N'}$ is an increasing function of $N'$ where $P_{1}=p_{i_{best}}$ and $P_{N}=1$.
Finally, given that we used different $N$ values for each dataset, we normalize $N^*$ by dividing it by the corresponding $N$.
Hence, instead of $N^*$, we systematically use $r = N^*/N$ in the following results.



To obtain a global view, we average the $r$ ratio across datasets and metric for each model and each method group and show the aggregation results in \cref{avg_critical_n}.
Note that we do note compute $r$ on all post-hoc methods at the same time because of the excessive computation cost. 
The modifications proposed improve the minimum benchmark size for every method group especially on the Class-Map methods where benchmark size can only be reduced up to $47.3\%$ with the baseline setting, whereas with the FP+FL setting, the size can be further reduced to $28.5\%$ of the original size on average.
However, consistently with \cref{results}, the effect of the proposed training settings depends on the method group considered.

\begin{table}[]
    \centering
    \resizebox{0.75\textwidth}{!}{%
    \begin{tabular}{c|cccc}\\ 
        \toprule 
        &Recent&Class-Map&BP&Perturbations\\ 
        \midrule 
        Baseline&81.1&47.3&25.8&70.0\\ 
        \hline 
        FP&80.9&32.9&28.0&82.8\\ 
        AP&79.4&56.0&36.4&55.6\\ 
        FL&83.8&51.8&26.7&67.1\\ 
        \hline 
        FP+FL&75.5&\textbf{28.5}&\textbf{22.7}&75.6\\ 
        AP+FL&77.1&60.1&34.1&\textbf{54.8}\\ 
        FP+AP&\textbf{72.4}&53.4&40.2&73.4\\ 
        \hline 
        FP+AP+FL&77.4&52.6&37.1&64.1\\ 
        \bottomrule 
        \end{tabular}
    }
    \caption{Minimum relative benchmark size $r$ for each model and post-hoc group.\label{avg_critical_n}}
\end{table}


\subsubsection{Model evaluation}

In this section we verify that the proposed modifications to the baseline training setting are having the indented effect.
\Cref{accuracy,calib} respectively shows the test accuracy and calibration measured with the AdaECE metric~\cite{adaece} on unmodified images (in the ''Regular`` columns) and on images perturbed by the faithfulness metrics (the 'FP' column).
Unsurprisingly, the training settings with FP yield models with superior accuracy on FP images compared to settings without FP, which confirms the value of training with FP batches. 
Also, the best-calibrated models are obtained following training using the FL (except on FP images on the time-lapse embryo dataset), showing the benefits of FL.
Finally, applying an adverarial attack on images with FP (i.e. FP+AP) helps to improve the calibration on the FP images on most datasets, especially on the time-lapse dataset (respectively $28.1$ and $5.6$ for the baseline and FP+AP settings). 

\begin{table}
    \centering
    \resizebox{\textwidth}{!}{%
    \begin{tabular}{l|cc|cc|cc|cc|cc} 
        \toprule \\ 
        Training setting & \multicolumn{2}{c}{CUB-200-2011} & \multicolumn{2}{c}{FGVC-Aircraft} & \multicolumn{2}{c}{Standford cars} & \multicolumn{2}{c}{CROHNIPI} & \multicolumn{2}{c}{Time-lapse embryos}\\ 
         & Regular & FP  & Regular & FP  & Regular & FP  & Regular & FP  & Regular & FP \\
        \midrule 
        Baseline & 2.6 & 8.3 & 6.2 & 15.9 & 2.3 & 9.1 & 10.1 & 17.7 & 18.2 & 28.1 \\ 
        FP & 8.4 & 8.2 & 9.3 & 6.7 & 4.3 & 4.2 & 8.2 & 7.8 & 11.4 & 9.1 \\ 
        AP & 7.5 & 17.3 & 10.5 & 28.8 & 3.5 & 15.6 & 5.3 & 17.8 & 1.9 & 8.4 \\ 
        FL & 4.0 & 9.6 & \textbf{1.8} & 14.9 & 1.3 & 7.0 & \textbf{3.2} & \textbf{5.9} & 10.1 & 18.1 \\ 
        \hline 
        FP+FL & 4.1 & \textbf{3.5} & 3.3 & \textbf{3.9} & 1.3 & \textbf{1.5} & 4.3 & 6.1 & 7.9 & 5.7 \\ 
        AP+FL & \textbf{2.6} & 10.1 & 4.4 & 18.8 & \textbf{1.0} & 10.1 & 5.2 & 12.5 & \textbf{1.5} & 12.6 \\ 
        FP+AP & 9.5 & 8.8 & 10.3 & 10.5 & 4.0 & 3.0 & 4.5 & 12.8 & 9.6 & \textbf{5.6} \\ 
        \hline 
        FP+AP+FL & 4.8 & 3.6 & 6.1 & 4.0 & 1.0 & 2.7 & 6.4 & 9.7 & 8.8 & 6.2 \\ 
        \bottomrule 
    \end{tabular}
    }
    \caption{Calibration error (AdaECE) of the models trained with the 8 studied training settings. Calibration is measured on unmodified images (in the ''Regular`` columns) and on images perturbed by the faithfulness metrics (the 'FP' column) \label{calib}}
\end{table}


\begin{table}
    \centering
    \resizebox{\textwidth}{!}{%
    \begin{tabular}{l|cc|cc|cc|cc|cc} 
        \toprule \\ 
        Training setting & \multicolumn{2}{c}{CUB-200-2011} & \multicolumn{2}{c}{FGVC-Aircraft} & \multicolumn{2}{c}{Standford cars} & \multicolumn{2}{c}{CROHNIPI} & \multicolumn{2}{c}{Time-lapse embryos}\\ 
         & Regular & FP  & Regular & FP  & Regular & FP  & Regular & FP  & Regular & FP \\ 
        \midrule
        Baseline & 76.6 & 44.1 & 79.7 & 48.0 & 87.1 & 56.4 & \textbf{83.9} & 73.3 & 63.8 & 40.6 \\ 
        FP & 74.1 & 57.2 & 78.0 & \textbf{62.9} & 87.5 & \textbf{69.5} & 81.1 & 72.2 & \textbf{65.4} & 54.7 \\ 
        AP & 72.7 & 37.0 & 74.3 & 36.5 & 84.0 & 44.5 & 83.5 & 70.4 & 62.3 & 36.4 \\ 
        FL & \textbf{79.0} & 43.0 & \textbf{79.8} & 43.8 & \textbf{87.6} & 52.6 & 83.5 & 66.0 & 63.2 & 42.7 \\ 
        \hline 
        FP+FL & 73.9 & \textbf{58.1} & 78.5 & 62.2 & 86.2 & 69.0 & 83.4 & \textbf{80.6} & 64.5 & \textbf{54.8} \\ 
        AP+FL & 72.0 & 37.3 & 75.4 & 36.4 & 84.1 & 46.0 & 83.2 & 68.1 & 60.5 & 30.5 \\ 
        FP+AP & 71.3 & 47.7 & 76.8 & 57.6 & 86.3 & 66.3 & 83.5 & 73.5 & 64.0 & 51.7 \\ 
        \hline 
        FP+AP+FL & 70.0 & 47.1 & 76.3 & 55.1 & 85.3 & 64.5 & 81.9 & 70.6 & 61.4 & 44.2 \\ 
        \bottomrule 
        \end{tabular} 
    }
    \caption{Accuracy of the models trained with the 8 studied training settings. Accuracy is measured on unmodified images (in the ''Regular`` columns) and on images perturbed by the faithfulness metrics (the 'FP' column)\label{accuracy}}
\end{table}

\subsubsection{Interpolation}

\begin{figure}[ht]
    \centering
    \includegraphics[width=\textwidth]{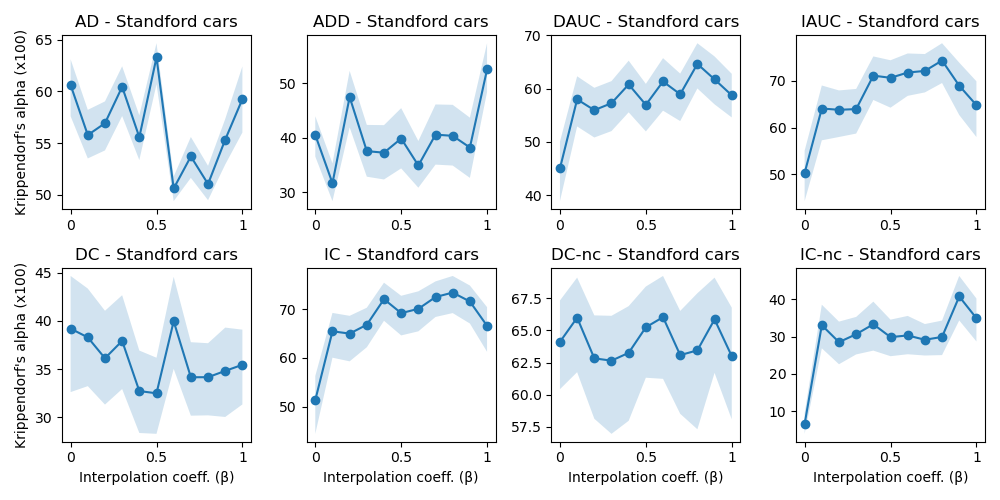}
    \caption{Krippendorf's alpha as a function of $\beta$ on the cars dataset. $\beta=0$ and $\beta=1$ respectively represents the baseline and the perturbed loss settings. Blue line indicates the mean value and the light blue shade indicates the $95\%$ confidence interval. Krippendorf's alpha values are multiplied by 100 for clarity.}
    \label{fig:alpha_interp}
\end{figure}

To further understand the impact of the FP on benchmark reliability, we trained models by interpolating between the regular setting and the FP setting. 
More precisely, we trained models with a loss defined as follows:

\begin{math}
    \mathcal{L}_{interp} =(2-\beta)\mathcal{L}_{CE}(I,y_c) + \beta \mathcal{L}_{perturb}(I^{FP},y_c),
\end{math}
with $\beta$ going from 0 to 1.
Note that when $\beta=1$, the loss is the same as the perturbed setting, but when $\beta=0$, the loss is scaled up compared to the baseline setting. 
We do this to ensure the global loss value is similar at every step of the interpolation.
Results in \cref{fig:alpha_interp} show that for the ADD, DAUC, IAUC, IC and IC-NC there is a clear trend upward when $\beta$ is increased, i.e., when the training setting become close from the perturbed loss setting. 
With the IAUC, IC and IC-NC metrics, $\alpha$'s value is already close from its final value ($\beta=1$) when $\beta=0.1$.
On the other hand, with the ADD and DAUC metrics, we observe regular increase of $\alpha$'s value until reaching $\beta=1$.
One explanation could be that IAUC, IC and IC-NC all perturb the image by applying a blur filter, which leaves out the low frequency information of the image. Hence, blurring is less destructive than replacing with black patches.
Given that the image is less altered and that information about the original content is still available, it could be easier for the model to fit the perturbed images.

\section{Discussion}

In this study, we proposed a method to improve the reliability of faithfulness metrics when benchmarking post-hoc explanation methods. 
By incorporating perturbed batches and by using an adaptive FL during model training, we observed significant enhancements in benchmark reliability, as quantified by Krippendorf's alpha. 

However, we acknowledge that this approach comes with an additional computation cost.
Computation time increases linearly with the number of batches processed at each step during training.
For example, a model trained in the FP+AP setting has to process the regular batch, the FP batch, and the AP batch.
Despite this additional cost, it is crucial to note that existing approaches, which apply post-hoc methods on standardly trained models, face considerable issues in benchmarking explanations from a faithfulness perspective. 
Indeed, faithfulness metrics have shown largely inconsistent rankings when applied to models trained under standard settings, rendering benchmarking nearly impossible without addressing this issue~\cite{sanity_checks}.

Future work could consist in understanding how the modifications proposed interact with each other and how the final rank reliability can be improved.
For example, one should study whether the AP are redundant with the FP given that some studies have shown that AP improves OOD generalization~\cite{adv_ood,adv_ood2,adv_ood3}.

In this work we used Krippendorf's alpha to measure the post-hoc methods benchmark consistency across images, following the work of Tomsett et al.~\cite{sanity_checks}.
However, Tomsett et al. also studied the reliability of faithfulness metrics by measuring the linear correlation between the faithfulness scores which allow to evaluate the ranking consistency across metrics and post-hoc methods~\cite{sanity_checks}.
Hence, it is essential that future research also include improving benchmark reliability as measured by the correlation between fidelity scores.


\section{Conclusion}

This paper presented several simple modifications to model training to enhance the benchmark reliability of post-hoc explanation methods in image classification, with a focus on faithfulness metrics. 
We tackled the issue of saliency metrics producing out-of-distribution samples by incorporating adversarial perturbations and perturbations resembling those encountered during the computation of faithfulness metrics into the training batches. Additionally, we trained the model using an adaptive Federated Learning (FL) approach instead of the standard Cross-Entropy (CE) loss.
Leveraging Krippendorf's alpha, a statistic from psychometrics, we quantifed the agreement among different saliency metrics, and showed that these training modifications significantly improves benchmark reliability on various datasets and faitfhulness metrics. 
To the best of our knowledge, this study is the first in the literature to tackle the challenge of improving the reliability of post-hoc faithfulness benchmarks. 
In the future, we plan to study benchmark reliability improvements using other measures than Krippendorf's alpha like linear correlation.
In conclusion, our research lays the foundation for more reliable evaluation practices, driving progress in post-hoc explanation methods and facilitating better interpretability of deep neural networks.


\bibliographystyle{plain} 
\bibliography{references} 

\begin{thebibliography}{10}

\bibitem{AD}
A.~{Chattopadhay}, A.~{Sarkar}, P.~{Howlader}, and V.~N. {Balasubramanian}.
\newblock Grad-cam++\: Generalized gradient-based visual explanations for deep
  convolutional networks.
\newblock In {\em 2018 IEEE Winter Conference on Applications of Computer
  Vision (WACV)}, pages 839--847, 2018.

\bibitem{pytorchcam}
Jacob Gildenblat and contributors.
\newblock Pytorch library for cam methods.
\newblock \url{https://github.com/jacobgil/pytorch-grad-cam}, 2021.

\bibitem{embryo_dataset}
Tristan Gomez, Magalie Feyeux, Justine Boulant, Nicolas Normand, Laurent David,
  Perrine Paul-Gilloteaux, Thomas Fréour, and Harold Mouchère.
\newblock A time-lapse embryo dataset for morphokinetic parameter prediction.
\newblock {\em Data in Brief}, 42:108258, 2022.

\bibitem{DC}
Tristan Gomez, Thomas Fr{\'e}our, and Harold Mouch{\`e}re.
\newblock Metrics for saliency map evaluation of deep learning explanation
  methods.
\newblock In Moun{\^i}m El~Yacoubi, Eric Granger, Pong~Chi Yuen, Umapada Pal,
  and Nicole Vincent, editors, {\em Pattern Recognition and Artificial
  Intelligence}, pages 84--95, Cham, 2022. Springer International Publishing.

\bibitem{tutorial}
Tristan Gomez and Harold Mouch{\`e}re.
\newblock {Computing and evaluating saliency maps for image classification: a
  tutorial}.
\newblock {\em {Journal of Electronic Imaging}}, 32(02), March 2023.

\bibitem{krippendorf_impl}
Thomas Grill and Santiago Castro.
\newblock Python implementation of krippendorff's alpha – inter-rater
  reliability.
\newblock \url{https://github.com/grrrr/krippendorff-alpha/}, 2023.

\bibitem{ADD}
Hyungsik Jung and Youngrock Oh.
\newblock Towards better explanations of class activation mapping.
\newblock In {\em Proceedings of the IEEE/CVF International Conference on
  Computer Vision (ICCV)}, pages 1336--1344, October 2021.

\bibitem{captum}
Narine Kokhlikyan, Vivek Miglani, Miguel Martin, Edward Wang, Bilal Alsallakh,
  Jonathan Reynolds, Alexander Melnikov, Natalia Kliushkina, Carlos Araya, Siqi
  Yan, and Orion Reblitz-Richardson.
\newblock Captum: A unified and generic model interpretability library for
  pytorch, 2020.

\bibitem{stanfordCars}
Jonathan Krause, Michael Stark, Jia Deng, and Li~Fei-Fei.
\newblock 3d object representations for fine-grained categorization.
\newblock In {\em 4th International IEEE Workshop on 3D Representation and
  Recognition (3dRR-13)}, Sydney, Australia, 2013.

\bibitem{Krippendorff}
Klaus Krippendorff.
\newblock {\em Content analysis: An introduction to its methodology}.
\newblock SAGE Publications, Inc., 2455 Teller Road, Thousand Oaks California
  91320, 2019.

\bibitem{aircraft}
S.~Maji, J.~Kannala, E.~Rahtu, M.~Blaschko, and A.~Vedaldi.
\newblock Fine-grained visual classification of aircraft.
\newblock Technical report, 2013.

\bibitem{focal_loss_calib}
Jishnu Mukhoti, Viveka Kulharia, Amartya Sanyal, Stuart Golodetz, Philip H.~S.
  Torr, and Puneet~K. Dokania.
\newblock Calibrating deep neural networks using focal loss.
\newblock In {\em Proceedings of the 34th International Conference on Neural
  Information Processing Systems}, NIPS'20, Red Hook, NY, USA, 2020. Curran
  Associates Inc.

\bibitem{adaece}
Jeremy Nixon, Michael~W. Dusenberry, Linchuan Zhang, Ghassen Jerfel, and Dustin
  Tran.
\newblock Measuring calibration in deep learning.
\newblock In {\em Proceedings of the IEEE/CVF Conference on Computer Vision and
  Pattern Recognition (CVPR) Workshops}, June 2019.

\bibitem{DAUC}
Vitali Petsiuk, Abir Das, and Kate Saenko.
\newblock Rise\: Randomized input sampling for explanation of black-box models,
  2018.

\bibitem{sanity_checks}
Richard Tomsett, Dan Harborne, Supriyo Chakraborty, Prudhvi Gurram, and Alun
  Preece.
\newblock Sanity checks for saliency metrics.
\newblock {\em Proceedings of the AAAI Conference on Artificial Intelligence},
  34(04):6021--6029, Apr. 2020.

\bibitem{crohnipi}
R{\'e}mi Vall{\'e}e, Astrid de~Maissin, Antoine Coutrot, Harold Mouch{\`e}re,
  Arnaud Bourreille, and Nicolas Normand.
\newblock {CrohnIPI: An endoscopic image database for the evaluation of
  automatic Crohn's disease lesions recognition algorithms}.
\newblock In {\em {SPIE Medical Imaging}}, volume 11317 of {\em Proc. SPIE,
  Medical Imaging 2020: Biomedical Applications in Molecular, Structural, and
  Functional Imaging}, page~61, Houston, France, February 2020. {SPIE}.

\bibitem{adv_ood3}
Riccardo Volpi, Hongseok Namkoong, Ozan Sener, John Duchi, Vittorio Murino, and
  Silvio Savarese.
\newblock Generalizing to unseen domains via adversarial data augmentation.
\newblock In {\em Proceedings of the 32nd International Conference on Neural
  Information Processing Systems}, NIPS'18, page 5339–5349, Red Hook, NY,
  USA, 2018. Curran Associates Inc.

\bibitem{CUB}
C.~Wah, S.~Branson, P.~Welinder, P.~Perona, and S.~Belongie.
\newblock {The Caltech-UCSD Birds-200-2011 Dataset}.
\newblock Technical Report CNS-TR-2011-001, California Institute of Technology,
  2011.

\bibitem{adv_ood2}
Qixun Wang, Yifei Wang, Hong Zhu, and Yisen Wang.
\newblock Improving out-of-distribution generalization by adversarial training
  with structured priors.
\newblock In S.~Koyejo, S.~Mohamed, A.~Agarwal, D.~Belgrave, K.~Cho, and A.~Oh,
  editors, {\em Advances in Neural Information Processing Systems}, volume~35,
  pages 27140--27152. Curran Associates, Inc., 2022.

\bibitem{adv_ood}
Mingyang Yi, Lu~Hou, Jiacheng Sun, Lifeng Shang, Xin Jiang, Qun Liu, and
  Zhiming Ma.
\newblock Improved ood generalization via adversarial training and pretraing.
\newblock In Marina Meila and Tong Zhang, editors, {\em Proceedings of the 38th
  International Conference on Machine Learning}, volume 139 of {\em Proceedings
  of Machine Learning Research}, pages 11987--11997. PMLR, 18--24 Jul 2021.

\end{thebibliography}

\end{document}